\documentclass{CVIS}
\usepackage{amsmath,bm}
\usepackage{graphicx}
\usepackage{dcolumn}
\usepackage{subcaption,soul}
\usepackage{caption}
\usepackage[numbers,sort&compress,sectionbib]{natbib}
\usepackage[pagebackref=flase,breaklinks=true,letterpaper=true,colorlinks=true,linkcolor = blue, urlcolor = black,citecolor=black,bookmarks=true]{hyperref}
\usepackage{url}
\usepackage{mleftright}
\usepackage{multirow}

\usepackage{titlesec}
\usepackage[redeflists]{IEEEtrantools}

\title{Machine Learning Challenges of Biological Factors in Insect Image Data}

\begin{document}
\bstctlcite{IEEEexample:BSTcontrol}
\sloppy
\author{
\begin{tabularx}{\textwidth}{X r}
Nicholas Pellegrino & Vision and Image Processing Group, System Design Engineering, University of Waterloo\\
Zahra Gharaee & Vision and Image Processing Group, System Design Engineering, University of Waterloo\\
Paul Fieguth & Vision and Image Processing Group, System Design Engineering, University of Waterloo\\
\multicolumn{2}{l}{Email: \{npellegr, zgharaee,  pfieguth\}@uwaterloo.ca}
\end{tabularx}
}

\maketitle
\begin{abstract} 
	The BIOSCAN project, led by the \textit{International Barcode of Life Consortium}, seeks to study changes in biodiversity on a global scale. 
	One component of the project is focused on studying the species interaction and dynamics of all insects.
	In addition to genetically barcoding insects, over 1.5 \emph{million} images per \emph{year} will be collected, each needing taxonomic classification. 
	With the immense volume of incoming images, relying solely on expert taxonomists to label the images would be impossible; however, artificial intelligence and computer vision technology may offer a viable high-throughput solution.
	Additional tasks including manually weighing individual insects to determine biomass, remain tedious and costly. 
	Here again, computer vision may offer an efficient and compelling alternative. 
	While the use of computer vision methods is appealing for addressing these problems, significant challenges resulting from biological factors present themselves. These challenges are formulated in the context of machine learning in this paper.
\end{abstract}

\section{Introduction}

The BIOSCAN project \cite{bioscan}, led by the \textit{International Barcode of Life Consortium} (iBOL) at the University of Guelph, is a large-scale, multinational research project with three main areas of study and corresponding objectives:
\begin{enumerate}
    \item \textbf{Species Discovery}: Generate genetic barcode \cite{hebert2003biological} coverage for \emph{two million} species.
    \item \textbf{Species Interactions}: Reveal species \emph{interactions} by targeting the symbiome.
    \item \textbf{Species Dynamics}: Monitor \emph{biodiversity} over time at 2,000 sites.
\end{enumerate}
To achieve this, at least 10 million life-specimens will be collected throughout the course of this project.
One particular component of the project focuses on studying insects, whereby insects from around the world will be both genetically barcoded and imaged.
In fact, the plan is to collect over 1.5 \emph{million} high-resolution images per \emph{year}, with each one needing taxonomic classification. 
With the immense volume of incoming images, relying solely on expert taxonomists to label the images would be impossible; however, artificial intelligence (AI) and computer vision (CV) technology may offer a viable high-throughput solution.

\begin{figure}[t]
    \centering
    \includegraphics[width=\columnwidth]{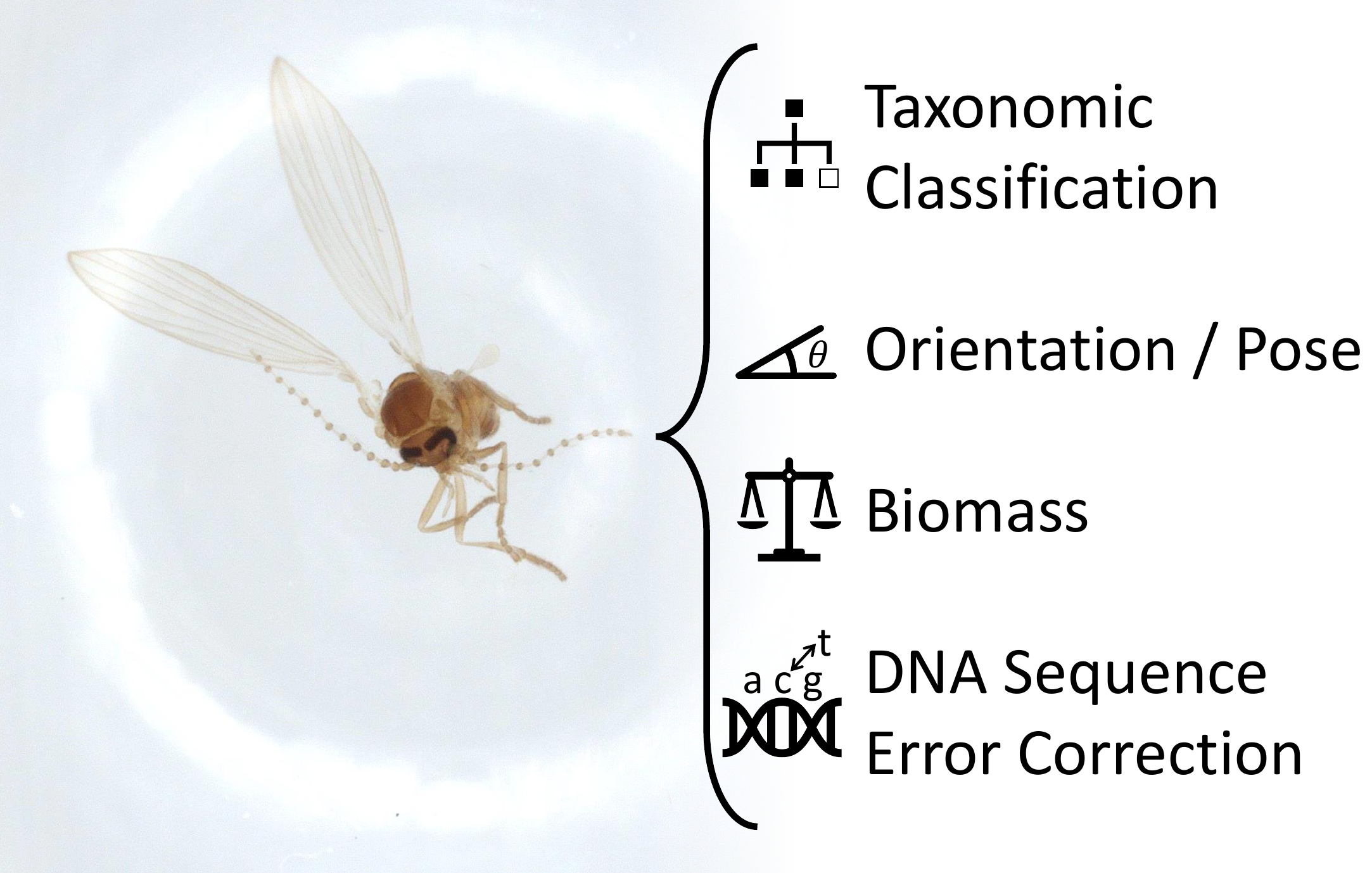}
    \caption{Valuable information may be inferred from images of single insects. AI methods have the potential to automatically extract such information. Insect imaged by the Centre for Biodiversity Genomics.}
    \label{fig:graphical_abstract}
\end{figure}

\begin{figure}[t]
    \centering
    \includegraphics[width=\columnwidth]{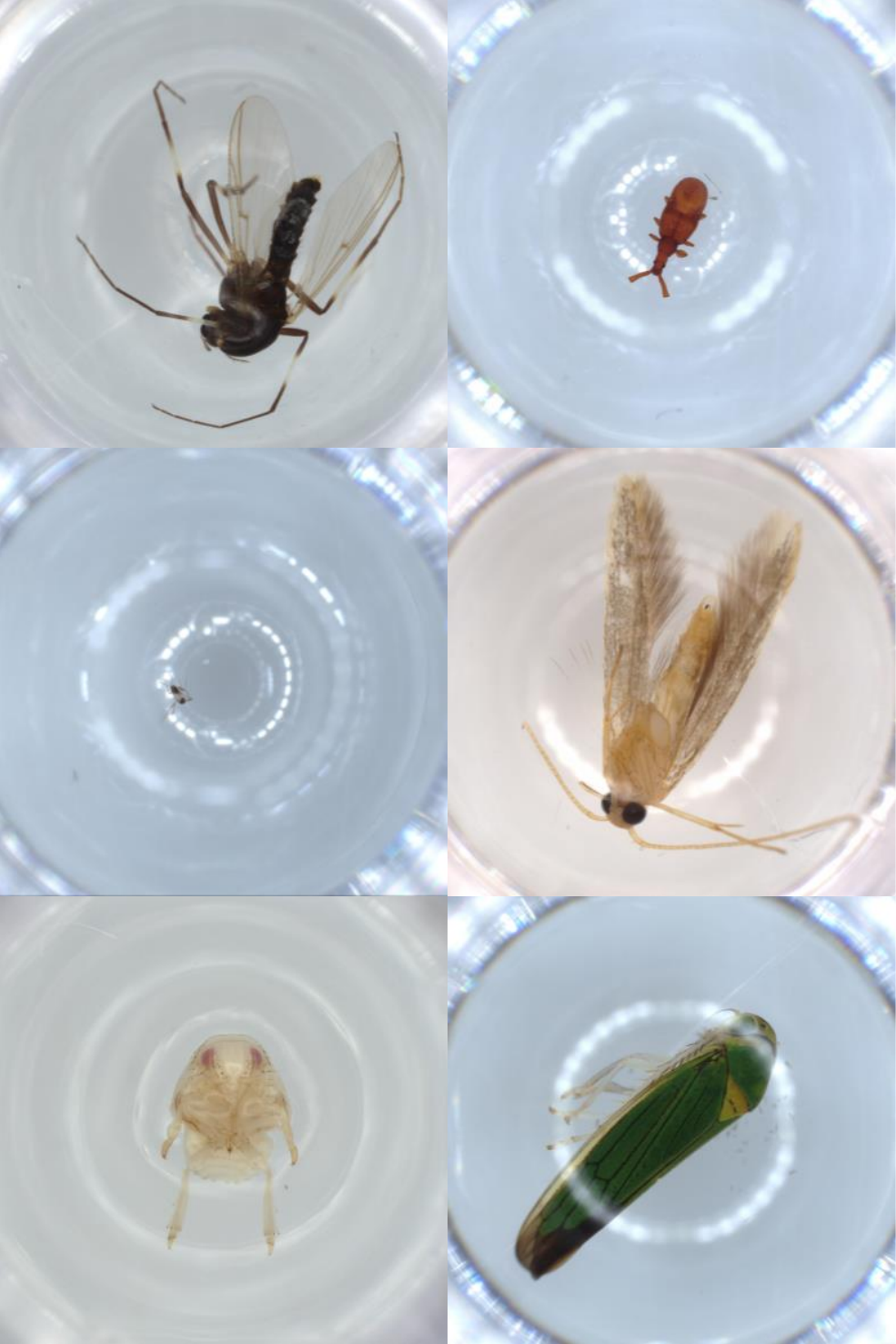}
    \caption{Composite of six insect images included in the data set. Observe the variety in insect size (scale), colour, transparency, orientation, pose, and illumination. Insects imaged by the Centre for Biodiversity Genomics.}
    \label{fig:insect_image}
\end{figure}

In addition to taxonomic classification, the use of CV / AI may enable further information such as insect biomass and insect orientation and pose to be inferred, useful for downstream tasks. Furthermore, it may even be possible to correct DNA sequencing errors where the measured sequence differs substantially from what is expected based on the image-based taxonomic assignment. 
Figure~\ref{fig:graphical_abstract} graphically illustrates the information we wish to extract / estimate from insect images to support the BIOSCAN project. 
The Centre for Biodiversity and Genomics, through BOLD \cite{bold_systems}, has provided a preliminary data set of images with included taxonomic annotations;
Figure~\ref{fig:insect_image} shows six example insect images from this data set. Across these images, large variation in insect size (scale), colour, transparency, orientation, pose and illumination is seen, behaviours that any estimator or information extraction tool would need to contend with.

\section{Proposed Scientific Outcomes}

This project is motivated by the opportunity for amazing scientific discoveries to be made that may be enabled through the effective application of machine learning and computer vision technology. This section briefly highlights several of such possible scientific outcomes.

\subsection{Biomass Estimation from Images}

Many reports world-wide indicate a grave decline in insect biomass \cite{hallmann2017more,hallmann2021insect} over long-term periods, where not only a significant loss of in insect life in bulk is projected to continue into the future, but a strongly correlated catastrophic drop in biodiversity (i.e., species loss). 

The use of insect traps (malaise traps) remains a common approach for monitoring insect abundance and biomass \cite{hallmann2017more,skvarla2021review}; however, the individual weighing of insects is still required.
By replacing the tedious task of weighing insects manually with the automatic process of computer-vision-based biomass estimation, the broader task of processing insects in the lab is streamlined, thus saving valuable time / resources of the BIOSCAN project for more impactful duties.

\subsection{Individual Species Identification through Bulk Insect Metabarcoding}

Bulk metabarcoding uses polymerase chain reaction (PCR) amplification technology to identify biological samples comprising hundreds or thousands of specimens \cite{taberlet2012towards,gibson2014simultaneous,lynggaard2019vertebrate}. In particular, this approach has been used for bulk barcoding of arthropods (the taxon containing insects). 

The availability of this method means that insect traps can be set up throughout forests and other ecosystems, collecting thousands of insects, and then through bulk processing, barcode information can (remarkably!) be produced for each \emph{individual} species present in the bulk sample.

The main reason for using this approach is to drive down the \emph{unit cost} of genetic barcoding. Previously, insects were carefully handled on an individual basis, meticulously being placed into separate test-tubes, before being genetically barcoded. The cost of labour to take on such an endeavour is enormous. However, by barcoding large groups of insects all at once, the cost is dramatically reduced and makes such a project all the more feasible.

Although bulk metabarcoding is already used today, the main contribution of the project in this area is the enormous \emph{amount} of samples that are being collected and which can efficiently be barcoded using this method.
    
\subsection{Species Interaction and Dynamics}

Beyond studying insects, iBOL seeks to barcode \emph{all} multicellular life. By doing so, it may become possible to, for example, sample an ecosystem by picking a leaf from a tree, extracting the DNA in bulk from all organisms on that leaf, and finally processing through barcode analysis to identify all (multicellular) life on that leaf. 
Such an approach would reveal the species interaction between the plant (host) and everything else living on it, perhaps ranging from lichens, algae, and fungi, to nematodes and perhaps even insects. 
Such a sampling and analysis technique offers far greater geographical resolution / precision than relatively non-discriminatory insect traps placed throughout biomes, allowing for the study of \emph{micro}-ecosystems.

Species interaction can be studied and made use of in other areas as well.
Many insects feed on or parasitize other larger plants or animals within their ecosystems and thus contain some DNA from those organisms within their digestive tracts / guts.
One such insect are mosquitoes, which regularly feed on animal blood. By catching and performing barcode analysis on mosquitoes found within a given ecosystem, much can be learned about the presence and perhaps even abundance of the animals from which they draw blood.
This sort of analysis may give valuable insights into species \emph{interactions} and, if repeated over spans of time, would illuminate details of species and ecosystem \emph{dynamics}.

\section{Machine Learning Challenges \& Directions}

\subsection{Limited Available Training Data for ML Tasks}
\label{subsec:lack_of_data}

Data annotation is both a costly and tedious process, necessary to create \emph{labelled} training data for downstream ML tasks. 
In particular, individually imaging, taxonomically classifying (by an expert), and weighing insects is an immense undertaking.
As a result, while iBOL has the ability to acquire millions of images annually, it \emph{does not} have the ability to label them at a similar rate while remaining cost-effective.
Gold standard data sets within the CV community, in particular object detection and recognition data sets, often include several million images, e.g., ImageNet \cite{deng2009imagenet} with $\sim$14M images, Open Images \cite{krasin2017openimages} with $\sim$9M images, and Microsoft Common Objects in Context (COCO) \cite{lin2014microsoftCOCO} with $\sim$2.5M images.
The great abundance of labelled data makes these data sets conducive to ML training tasks. 
While the data set currently available for this project is composed of roughly 1 million labelled images, many of these images are unusable as a result of processing issues whereby contaminants from other insects have compromised the associated barcode information, leaving closer to 200,000 vetted records.
As a result of the having relatively limited training data, conventional methods or network architectures, which are normally effective when vast amounts of training data are available, are likely to perform poorly.
Beyond training, the lack of labelled data creates a complication for model testing (i.e., determining whether the model is able to generalize to new, unseen data), since the limited data set may not adequately represent the variation present within the entire broad taxonomic class of insects. 

One might imagine augmenting the available data with labelled images from other data sets. In principle, this would help; however, to the best of the authors' knowledge, the largest available insect data set, IP102 \cite{wu2019ip102} contains only 75,000 images, and of which only 19,000 are annotated. Furthermore, the IP102 data set is quite limited in scope, containing only images of 102 insect species. Because of its relatively small size in comparison to the images already available through BOLD, differences in formatting, and image quality / staging, integrating the two data sets would be difficult while unfortunately providing little benefit.

\subsection{Intrinsic Challenges Associated with Insect Data}

Beyond the \emph{quantity} of data available, as was discussed in Section~\ref{subsec:lack_of_data}, several complicating factors, intrinsic to the data set, exist. These are listed here, each with some discussion.

\begin{figure}[t]
    \centering
    \includegraphics[width=\columnwidth,trim={0.2cm 0 1.2cm 0},clip]{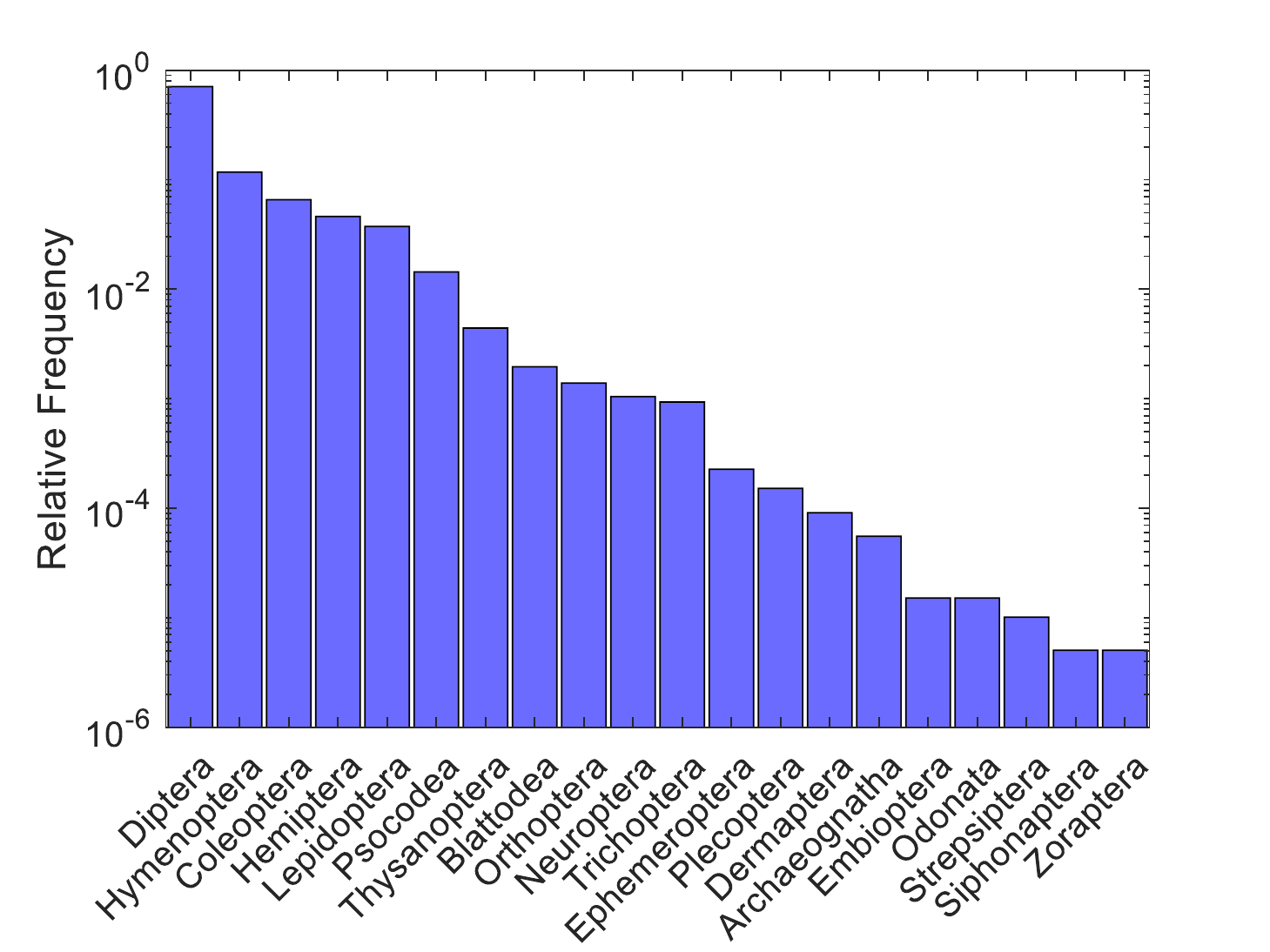}
    \caption{Histogram showing relative frequency of each insect taxonomic order present within the data set, forming a long-tailed distribution. Notice the log-scaled vertical axis. The order diptera (flies) accounts for roughly 70\% of all samples.}
    \label{fig:order_histogram}
\end{figure}

\begin{enumerate}
    \item \textbf{Class imbalance}:\\ An imbalance exists between classes, exhibiting a long-tailed distribution, characteristic of real-world biological data. This causes difficulty when training classification models because relatively few classes comprise the majority of the data set. 
    In fact, at the taxonomic order-level, the order diptera (flies) accounts for the majority of all available training data, comprising roughly 70\% of the entire data set. Figure~\ref{fig:order_histogram} shows the relative frequency distribution of samples at across all taxonomic orders present. The long-tailed distribution is quite apparent. 
    This degree of imbalance means that classifiers are able to achieve considerable accuracy ($\sim$70\%) on the \emph{available} data set simply by \emph{always} predicting diptera and without  actually \emph{learning} the characteristics of diptera vs.\ other orders \cite{harvie2022computer}! Furthermore, there are many taxonomic orders with \emph{very few} training images available, posing additional challenge.
    \item \textbf{Variation in insect Size}:\\ Insects that are especially small appear quite small in the images, whereas other larger insects may occupy the entire frame, as seen in Figure \ref{fig:insect_image}.  This means that the sorts of resolvable physical features of the insects vary considerably with their size. Furthermore, by virtue of occupying fewer image pixels, less information is actually captured in images of smaller insects. 
    \item \textbf{Visual similarity of distinct insect species}:\\ Distinct insect species may appear visually similar for a variety of reasons, most common of which is high genetic / taxonomic similarity. For example, many beetles appear visually similar and are characterized by traits such as their rounded shell, often covering a pair of slightly transparent wings. Other reasons for visual similarity between distinct species include mimicry / imitation \cite{rettenmeyer1970insect,van1958experimental,ritland1991viceroy}, or more broadly, convergent evolution \cite{maruyama2017deep}.
    \item \textbf{Visual differences based on pose and orientation of the insect samples}:\\ Insects exhibit bilateral symmetry, however, that symmetry is immediately broken in images where insects are not viewed parallel to their sagittal plane, a result of their \emph{orientation} in the image. Furthermore, this symmetry is broken by differences between left and right in the insects' \emph{pose}, whereby left and right limbs or joints may be at differing angles. 
    Beyond bodily symmetry, it is clear that insects appear differently when viewed from different angles (front vs.\ back, top vs.\ bottom, etc.) and when limbs are differently positioned. 
    \item \textbf{Metamorphosis and variation associated with age / stage within life cycle}:\\ Insects undergo enormous physical transformations as they transition through various stages of life, from eggs, to larvae or nymphs, to pupae, and finally adults. 
    Consider the order lepidoptera, consisting of butterflies and moths, which transform dramatically throughout their life --- transitioning from caterpillar, to chrysalis, and finally to adult butterfly/moth --– all the while remaining taxonomically identical!
    These changes throughout the life cycle of insects result in significant intra-class variation, but also may increase inter-class similarity, given that many insect larvae are ``worm-like'', and even worse, all insect eggs can be described as oblong elliptic structures, though considerable variations in size (8 orders of magnitude) and aspect ratio do exist \cite{church2019dataset}.
    \item \textbf{Sexual dimorphism}:\\ Sexual dimorphism, \emph{meaning two forms}, is the physical difference between taxonomically identical insects (i.e., same species) of opposite sex. Examples range from difference in overall size, to proportions (e.g., limbs or antennae of differing lengths), to colouration and patterning \cite{vendl2018hidden}, and even whether an insect poses certain body parts (e.g., cephalic and/or pronotal horns in male scarab beetles \cite{ahrens2014evolution}). As in the preceding example, such differences contribute to the intra-class variance.
\end{enumerate}
Many of these problematic biological challenges can be understood and succinctly summarized as factors that contribute to causing significant intra-class variance, while at the same time decreasing inter-class separation (i.e., increasing inter-class similarity). 
Such challenges often present themselves in the domain of fine-grained classification, a category of ML / CV problems whereby classes must be discriminated based on subtle or localized differences \cite{peng2017object,shroff2020focus}. 

\subsection{Possible Directions Forward}

Several potential and overlapping paths forward exist to address the challenges associated with the lack of data, high intra-class variance, and high inter-class similarity. These include the following.

\begin{enumerate}
    \item \textbf{Data augmentation and re-sampling}: \\Since \emph{some}, roughly 200,000, images do have labels already assigned to them, perhaps data augmentation \cite{van2001art,shorten2019survey} can be used to artificially inflate the size of the available training data set, thus enabling the use of conventional / tried-and-true CV methods. Random re-sampling \cite{japkowicz2001concept} is another approach that can be applied to simultaneously address the between-class and within-class imbalances problems.
    
    \item \textbf{Sparse models}: \\In the absence of a suitably large training data set, careful selection of DNN/CNN model architectures may be necessary. Specifically, sparse models --- those with relatively \emph{few} weights --- may offer a suitably generalized solution such that over-fitting to the small training data set can be avoided \cite{xu2019overfitting}. 
    
    \item \textbf{Alternative learning techniques}: \\A further alternative would be to employ techniques such as domain adaptation \cite{farahani2021brief,csurka2017comprehensive}, transfer learning \cite{pan2009survey,weiss2016survey}, domain generalization through feature representation \cite{muandet2013domain,li2018domain} or meta-learning \cite{li2018learning,balaji2018metareg}, perhaps allowing a network model with sufficient object --- possibly specifically \emph{animal} --- classification ability to be fine-tuned using the limited insect data set such that it is able to differentiate and adequately classify insects.
    
    \item \textbf{Alternative loss functions}:\\
    Certain loss functions are designed in ways that may mitigate the impact of various inadequacies in the training data. One such example is the Focal Loss \cite{lin2017focal}, which is designed specifically to address issues of class imbalance, and does so by weighting the loss associated with \emph{``difficult''} training examples --- those likely to be misclassified --- more than training samples that are ``easy'' or likely to be classified correctly.
\end{enumerate}

\section{Conclusions}
While numerous challenges prevent the simple application of existing computer vision techniques to assist in the BIOSCAN project, addressing these challenges would be a contribution to the ML / AI field, while also being highly meaningful to the broader international community of BIOSCAN collaborators.
Future works will involve implementations and evaluations of the the proposed ML strategies for overcoming the challenges formulated in this paper. 
Such contributions may enable the tracking of changes in insect abundance, biomass, and diversity at the species level over time and across a multitude of geographical locations. This information is critical to understanding the impacts of ecological change and for formulating strategies for mitigating further catastrophic damage to the global ecosystem.

\section*{Acknowledgments}
The authors thank the Centre for Biodiversity Genomics at the University of Guelph for access to the insect data within BOLD \cite{bold_systems}.

\bibliographystyle{IEEEtran}  
\bibliography{references}

\end{document}